\documentclass[a4paper,twoside]{article}

\usepackage{epsfig}
\usepackage{subfigure}
\usepackage{calc}
\usepackage{amssymb}
\usepackage{amstext}
\usepackage{amsmath}
\usepackage{amsthm}
\usepackage{multicol}
\usepackage{pslatex}
\usepackage{apalike}

\usepackage{float}
\usepackage{url}            
\usepackage{booktabs}       
\usepackage{amsfonts}       
\usepackage{nicefrac}       
\usepackage{microtype}      
\usepackage{siunitx}
\usepackage{caption} 
\usepackage[table,xcdraw]{xcolor}
\usepackage{spverbatim}
\usepackage{graphicx}
\usepackage{makecell}
\usepackage{graphicx} 
\usepackage{hyperref}
\usepackage{dblfloatfix}    
\usepackage{xcolor}
\hypersetup{
  colorlinks,
  citecolor=black,
  linkcolor=black,
  urlcolor=blue}

\graphicspath{ {images/} }  
\usepackage[inline]{enumitem}

\usepackage{SCITEPRESS}     

\begin{document}

\title{DeepGeo: Photo Localization with Deep Neural Network}

\author{\authorname{Sudharshan Suresh, Nathaniel Chodosh, Montiel Abello}
\affiliation{Robotics Institute, Carnegie Mellon University \\ Pittsburgh, PA 15213, USA }
\email{\{sudhars1, mabello, nchodosh\}@andrew.cmu.edu}
}


\abstract{In this paper we address the task of determining the geographical location of an image, a pertinent problem in learning and computer vision. This research was inspired from playing GeoGuessr, a game that tests a humans' ability to localize themselves using just images of their surroundings. In particular, we wish to investigate how geographical, ecological and man-made features generalize for random location prediction. This is framed as a classification problem: given images sampled from the USA, the most-probable state among 50 is predicted. Previous work uses models extensively trained on large, unfiltered online datasets that are primed towards specific locations. To this end, we create (and open-source) the \textbf{50States10K} dataset - with 0.5 million Google Street View images of the country. A deep neural network based on the ResNet architecture is trained, and four different strategies of incorporating low-level cardinality information are presented. This model achieves an accuracy 20 times better than chance on a test dataset, which rises to 71.87\% when taking the best of top-5 guesses. The network also beats human subjects in 4 out of 5 rounds of GeoGuessr.}

\onecolumn \maketitle \normalsize \vfill

\section{Introduction} \label{intro}
Games such as \href{https://geoguessr.com/}{GeoGuessr} demonstrate that humans are remarkably good at coarse estimates of location from a single image. The game displays a Google Street View \ang{360} panorama from anywhere in the world and asks the user to guess the GPS coordinates of the place the photo was captured. Humans can achieve ballpark estimates in spite of the fact that many (if not most) images have ambiguous locations unless they contain very specific landmarks. 

Natural images have numerous cues that allow inference: vegetation, and man-made infrastructure such as roads, marking and signs. Humans are also able to use other cues to find out where they are. Common examples of these are driving directions, languages and prior knowledge of vegetation and geography. We attempt to replicate this semantic reasoning using a deep neural network. The abundance of open-source visual data and previous success of learning methods in scene recognition motivates this effort. 

Previous works, based on both learning as well as scene matching methods, have shown appreciable results at various spatial scales. We restrict the scope of the classification problem to the USA. This is an ideal playground for such a method due to its near-perfect street view coverage, high road density and varied geography. Segmenting the USA into a set of $50$ classes based on state boundaries, we predict the state that an organized set of input images belongs to. In contrast to previous works that classified individual images, our goal is to train a network to be effective at geolocalization from panoramic viewpoints that are provided in GeoGuessr. Each input sample contains four images taken at the same location, oriented in the cardinal directions. Given such an input sample, we classify the scene to a one-hot vector label {\small$Y_i \in \mathbb{R}^{50}$}.

Automatic image localization is a challenging problem that tests the capabilities of deep neural networks. This tool will aid in wide-ranging applications such as metadata aggregation, location-based priming and the creation of richer datasets. We demonstrate that deep neural networks are capable of effective geolocalization without requiring a large dataset and exhaustive training.  Several approaches for integrating cardinality are employed to determine the most effective strategy for concatenation of learned features. 

\begin{figure*}[t]
 \centering
 \includegraphics[width=0.8\linewidth]{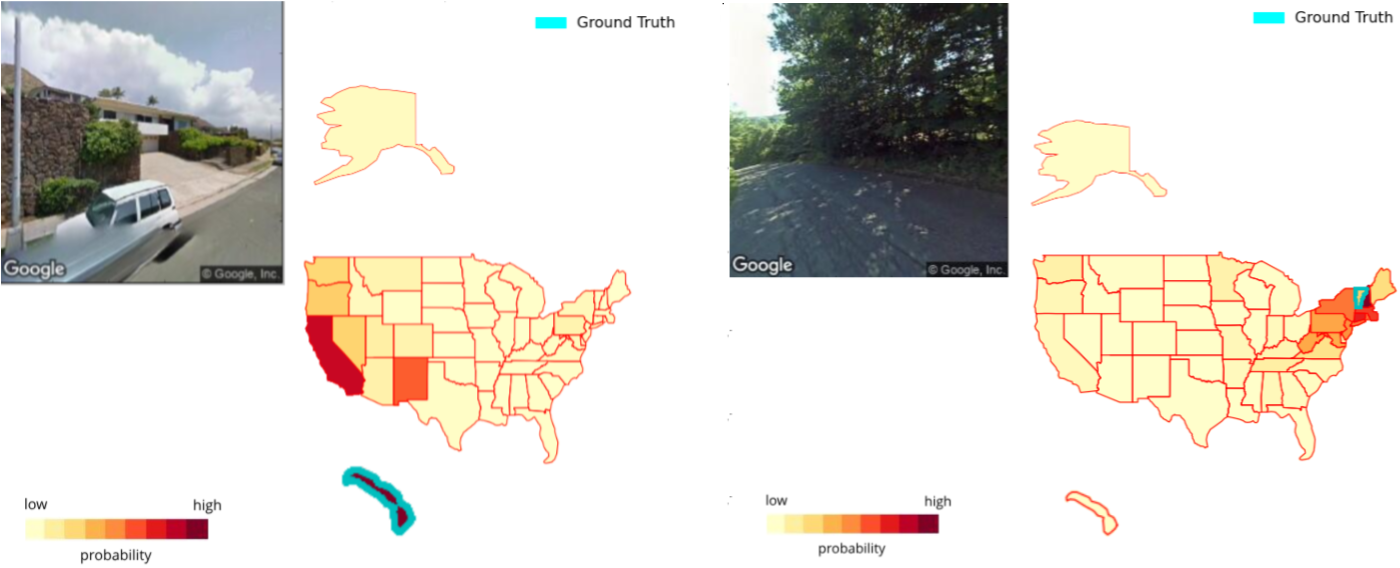}
 \caption{Visualization of the heat-maps of a test image taken from (a) Hawaii (b) Vermont  (best viewed on computer screen and in color). While the network correctly classifies (a) to Hawaii, it also indicates a high similarity score with California. In (b) one could argue that Vermont possesses characteristics shared by other parts of the Northeast, like upstate NY.}
 \label{fig:heatmaps}
\end{figure*}

Our key contributions are \textbf{(a)} DeepGeo\footnote{{\scriptsize  \url{https://github.com/suddhu/DeepGeo}}}: a deep neural network that predicts the most-probable state a given image was captured in (not restricted to cities/famous landmarks) \textbf{(b)} the \href{https://drive.google.com/file/d/1Y8eqx1Uy8kuRP4BNmTCVvrNbCxx6RoiP/view}{\textbf{50States10K}}\footnote{{\scriptsize \url{https://goo.gl/dM7xdk}}} dataset, a comprehensive street view dataset of the USA. 

\section{Related Work} \label{rel}
The challenges of geolocation based on visual information has its foundations in \cite{thompson1996geometric}. There have been various formulations of the same problem over the years that incorporate geotagged online photos \cite{hays2008im2gps,weyand2016planet,cao2013graph}, satellite aerial imagery \cite{arandjelovic2016netvlad} and city skylines \cite{ramalingam2010skyline2gps}. Due to their extensive coverage and accuracy, Google Street View scenes have been used in \cite{zamir2010accurate,zamir2014image,kim2015predicting}. However, these works only tackle localization at the city level, most using feature-based methods. For example \cite{zamir2014image} restricts geolocation to Pittsburgh and Orlando via SIFT descriptors and feature points. 

PlaNet \cite{weyand2016planet} and IM2GPS \cite{hays2008im2gps} are considered the baselines for scalable image geolocation over the entire world. IM2GPS computes the closest match via scene matching with a large corpus of 6 million geotagged Flickr images. They perform purely data-driven localization with features such as color and geometric information.

\begin{figure*}[t]
 \centering
 \includegraphics[width=\linewidth]{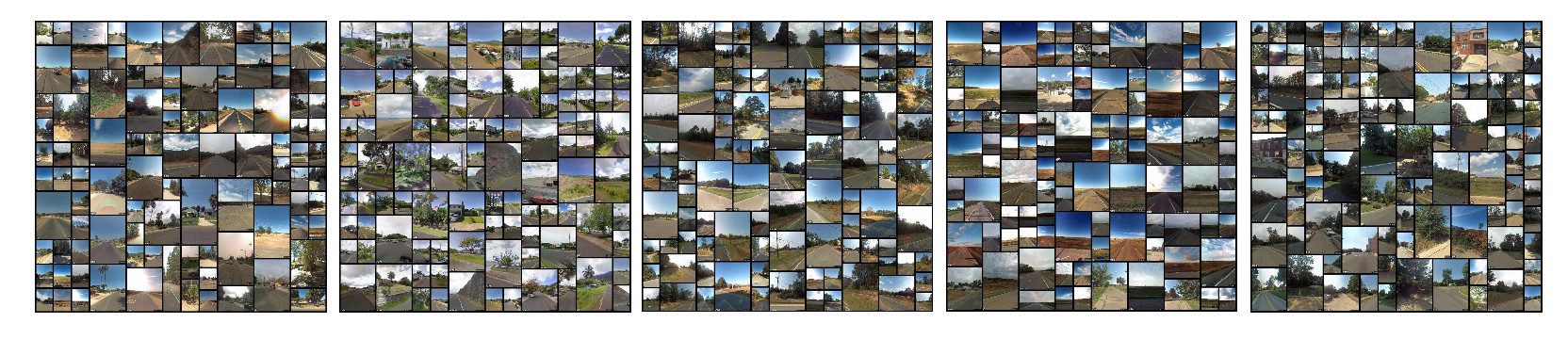}
 \caption{Sample images - California, Hawaii, North Carolina, North Dakota, and Pennsylvania datasets}
 \label{fig:collage}
\end{figure*}

PlaNet \cite{weyand2016planet} uses a larger such dataset paired with a deep network to obtain a probability distribution of location over a partitioned world map. It also represents the first attempt to harness a deep network to achieve coarse global localization. They formulate their task as classifying a test image to one of many partitioned multi-scale geographic cells. It uses a CNN-based on the Inception architecture, trained with 200 cores for 2.5 months. 

These methods rely on sheer volume of training images - a perusal of their data yields many indoor scenes, pictures of food and pets. They are also naturally primed towards famous landmarks and cities, and are not representative of \textit{throw a dart at a map} scenarios such as GeoGuessr. As mentioned in Section \ref{intro}, our method differs from the aforementioned works both in terms of classification goals and data collection. While PlaNet correctly classifies 28.4\% of the test data at the country level, IM2GPS has a test set median error of 500km. It is noted that these cannot be used as comparison metrics with our method, given they attempt to tackle the problem of \textit{global} localization while we wish to perform classification to a state in the USA. The above two methods also do not utilize Google Street View imagery to any extent. It is worth mentioning that PlaNet bested human subjects in GeoGuessr in 28 rounds of 50. 

While the above two methods use large Flickr datasets, \cite{zamir2010accurate} uses a dataset of 100,000 GPS-tagged Street View images in two cities. To the best of our knowledge, there are no existing large-scale Google Street View datasets of either the world or the USA.

\section{Dataset} \label{dataset}
\subsection{Location Sampling} \label{sampling}
\begin{figure}[H]
 \centering
 \includegraphics[width=\linewidth]{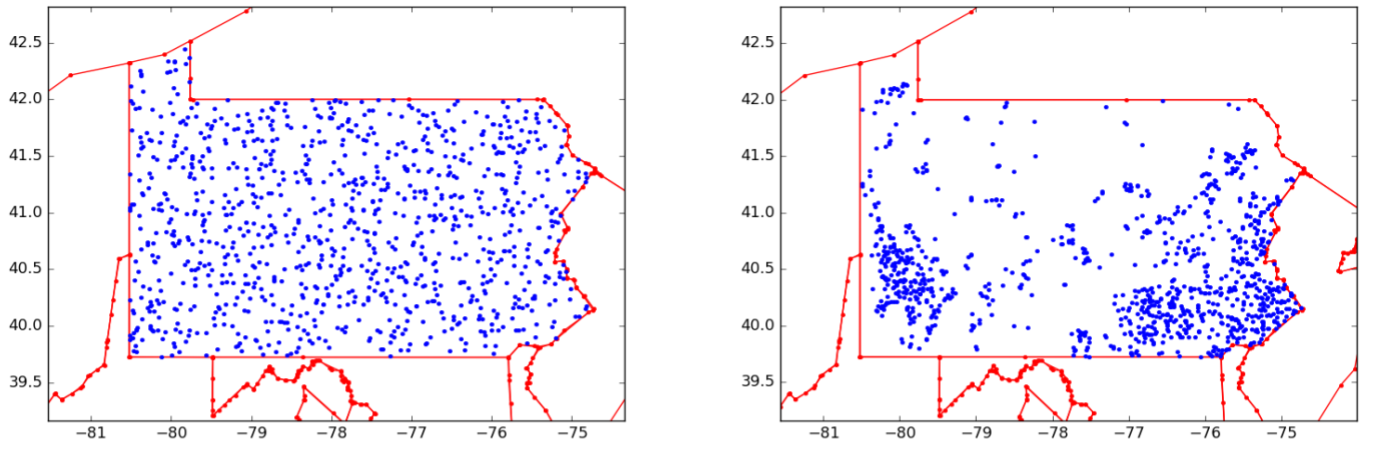}
 \caption{Sample locations in Pennsylvania for (i) uniform and (ii) our population density aware approaches. We see more images sampled from populated areas.}
 \label{fig:penn}
\end{figure}

Rather than sampling uniformly across the USA, we sample at an equal number of locations within each state to ensure that each class is properly represented. To sample locations within each state, we use state boundary data\footnote{{\scriptsize \url{http://econym.org.uk/gmap/states.xml}}}. State boundaries are represented by a polygon connecting an ordered set of points. We sample \texttt{[latitude, longitude]} uniformly within the minimum rectangle enclosing this polygon. Points falling outside the state boundary are discarded. 

The \textit{Gridded Population of the World v4} \footnote{{\scriptsize \url{http://sedac.ciesin.columbia.edu/data/collection/gpw-v4/documentation}}} is used to inform our location sampling. We use the population density map with a grid size of $2.5$ arc-minutes in latitude and longitude. Within each state we normalize the population density values by dividing each grid value by the maximum density in the state, and discard sampled locations with a population density below a threshold $p_{min}$. We also ensure that duplicate locations are not sampled to ensure good coverage.

\begin{figure*}[t]
 \centering
 \includegraphics[width=0.32\textwidth]{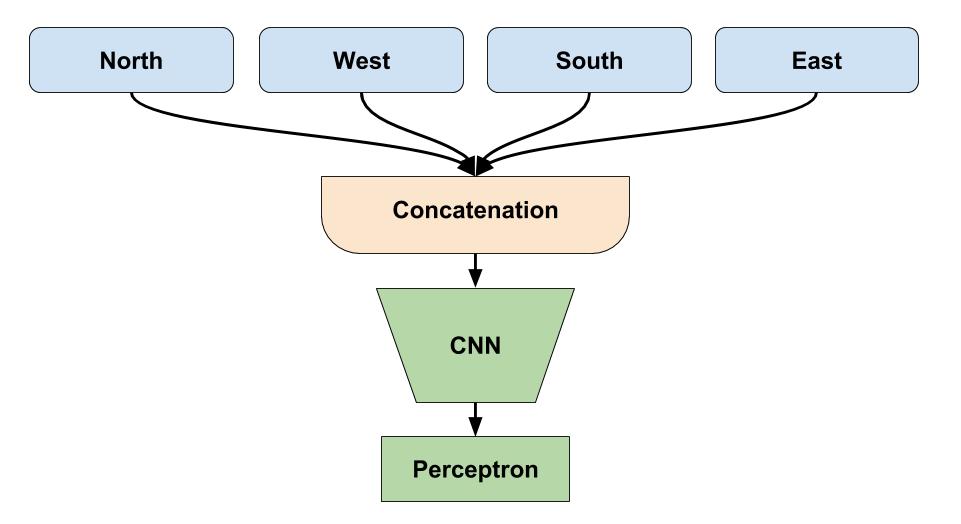}
 \includegraphics[width=0.32\textwidth]{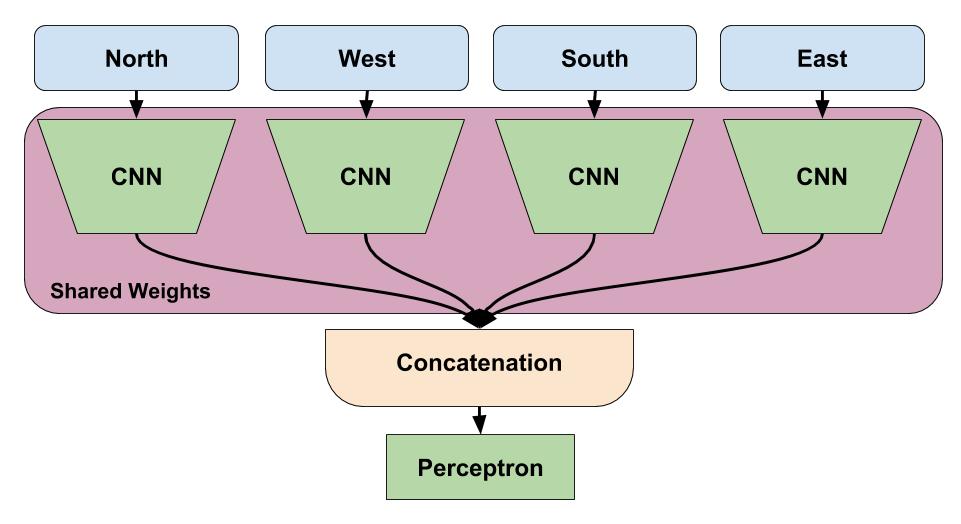}
 \includegraphics[width=0.32\textwidth]{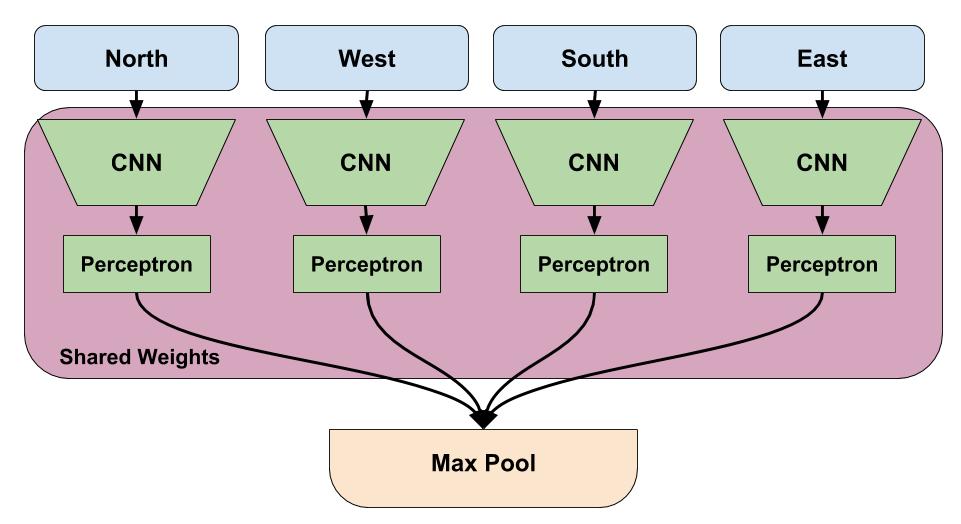}
 \caption{From left to right - Early, Medium and Late Integration. The networks are viewed from top to bottom.}
 \label{fig:networks}
 \end{figure*}

This density-based sampling allows us to achieve a good spread of locations, primes the model for more densely populated regions and boosts the hit rate of the image scraping tools described in section \ref{scraper}. For example figure \ref{fig:penn} shows population aware sampling in Pennsylvania. 

\subsection{Data Scraping} \label{scraper}
\begin{figure}[H]
 \centering
 \includegraphics[width=\linewidth]{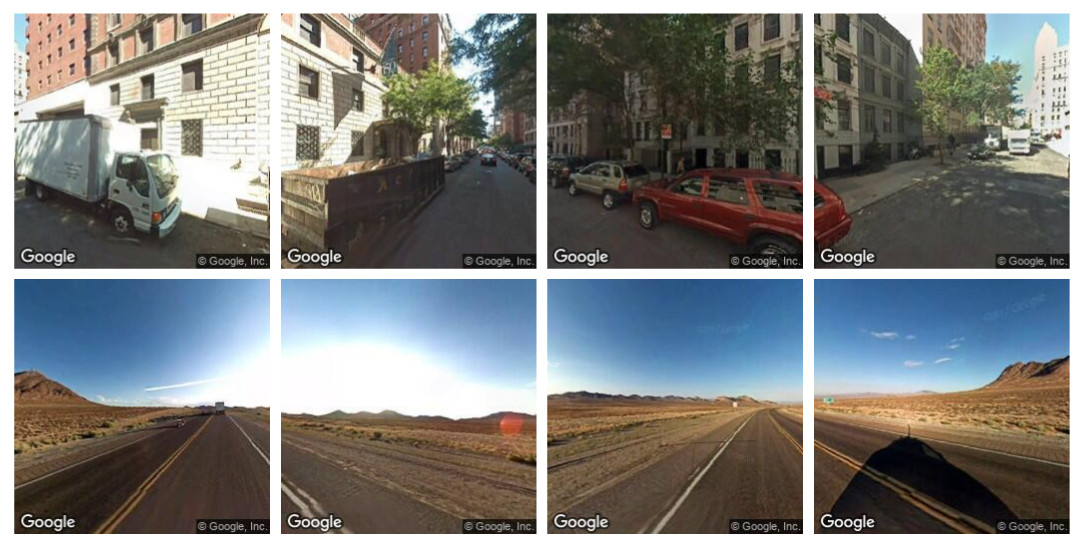}
 \caption{Two input samples consisting of four images each at headings of $[0^{\circ},90^{\circ},180^{\circ},270^{\circ}]$.}
 \label{fig:example}
\end{figure}

Out dataset is obtained via the Google Street View API which provides tools for image extraction. Images were gathered over a period of $5$ days, running our image scraping tools on several machines. 

The sampled locations from section \ref{sampling} are fed as queries to a script, which retrieves the closest unique panorama. These $360^{\circ}$ images are split into 4 rectified images facing the cardinal directions, as shown in Figure \ref{fig:example}. Duplicate images are pruned away to give $4 \times 2500 \times 50$ unique natural scene images.

\subsection{Dataset description}  \label{description}

We present the \textbf{50States10K} dataset. To the best of our knowledge, this is the first large-scale, open-source Google Street View dataset of the USA. There are $2500$ samples ($10000$ unique images) in each of the $50$ states to provide a total of $125000$ samples (0.5 million images). Each sample contains $4$ rectified, $256 \times 256$ Street View images taken from the sample location, facing the cardinal directions. The latitude and longitude of each sample are also provided. This dataset is used to train our classifier. Figure \ref{fig:collage} shows example images highlighting similarities and differences in features across several states.

For testing purposes, we created the \texttt{50States2K} dataset. There are $500$ samples ($2000$ unique images) in each state, giving a total of $25000$ samples. Test locations were sampled carefully to ensure uniqueness from one another and from locations in the training set.

We make the \href{https://drive.google.com/file/d/1Y8eqx1Uy8kuRP4BNmTCVvrNbCxx6RoiP/view}{\textbf{50States10K}}\footnote{{\scriptsize {\scriptsize \url{https://goo.gl/dM7xdk}}}}  and \href{https://drive.google.com/file/d/1zHc3wcQxicoij0tGHBpVSKqN_QgZsF-t/view}{\textbf{50States2K}}\footnote{{\scriptsize {\scriptsize \url{https://goo.gl/LddtEx}}}}  datasets available for open source use. Each set organizes images into folders based on state location and provides the latitude and longitude corresponding to each image.

\section{Methods}  \label{methods}
As a baseline method we performed standard CNN classification, treating each image as a separate training example. That is we did not group multiple views from the same location in any way. The Residual Network (ResNet) is a recently published CNN architecture which we determined to be suitable for our problem\cite{resnet}. The ResNet architecture uses so called \textit{residual blocks} to achieve better gradient propagation, which enables training much deeper networks. 

In \cite{resnet} the authors present variations of their network that have 18 to over 100 layers, we determined empirically that the 50-layer variant was sufficient. We then tested three variants of this network to determine the best way of integrating the information from multiple views of the same location. Some previous works have used sophisticated LSTM setups to accomplish this \cite{weyand2016planet}. However in this case each input has a fixed number of input images, making an LSTM overly complicated. Additionally the four input images have a fixed relationship, since they always correspond to views in cardinal directions. Thus, we chose to test simpler integration techniques described below and shown in Figure \ref{fig:networks}.
 
\begin{description}
    \item[\underline{Early Integration}]: In our early integration scheme all four views are concatenated to form a twelve channel  $( \{R,G,B\} \times  \{N,S,E,W\} )$ input image, allowing information to be shared between images in all layers of the network. The network is otherwise a standard 50-layer ResNet, the only modification being that the number of filters in each block of the network is doubled to account for the extra input channels.
    \item[\underline{Medium Integration}]: In the medium integration network we effective make the assumption that the feature extraction for each of the input images should be constant, and that integration should be done only on the extracted features. To this end, we use a CNN to get a low dimensional summary of each image, concatenate the features of each, and run a single layer perceptron on the new feature for classification. The CNN is shared across all four images, and is simply the 50-layer resnet with the fully connected layer removed.
    \item[\underline{Late Integration}]: For late integration we also run a shared CNN over each input image and then combine the results. However, in this scheme the CNN includes the fully connected layer and to integrate the predictions we simply take the maximum over each output class. The idea being that the only information to be gained by integration is being able to use the “best view” for the predictions, i.e the one that assigns the highest confidence to its prediction.
\end{description}
\subsection{Implementation Details}
We implemented the network in Tensorflow as described in \cite{resnet} and used the Adam algorithm for training with a learning rate of $0.001$,  $\beta_1=0.9$, $\beta_2=0.999$ and $\varepsilon = 10^{-8}$. We also applied weight decay with a weight of 0.0001. All three variants were trained for fifty epochs using a 90/10 train/validation split. The epoch which achieved the highest validation accuracy was then used on the test set. Training took approximately 30 hours per network running on a single Nvidia Titan X GPU.

\begin{figure*}[b]
 \centering
 \includegraphics[width=0.9\linewidth]{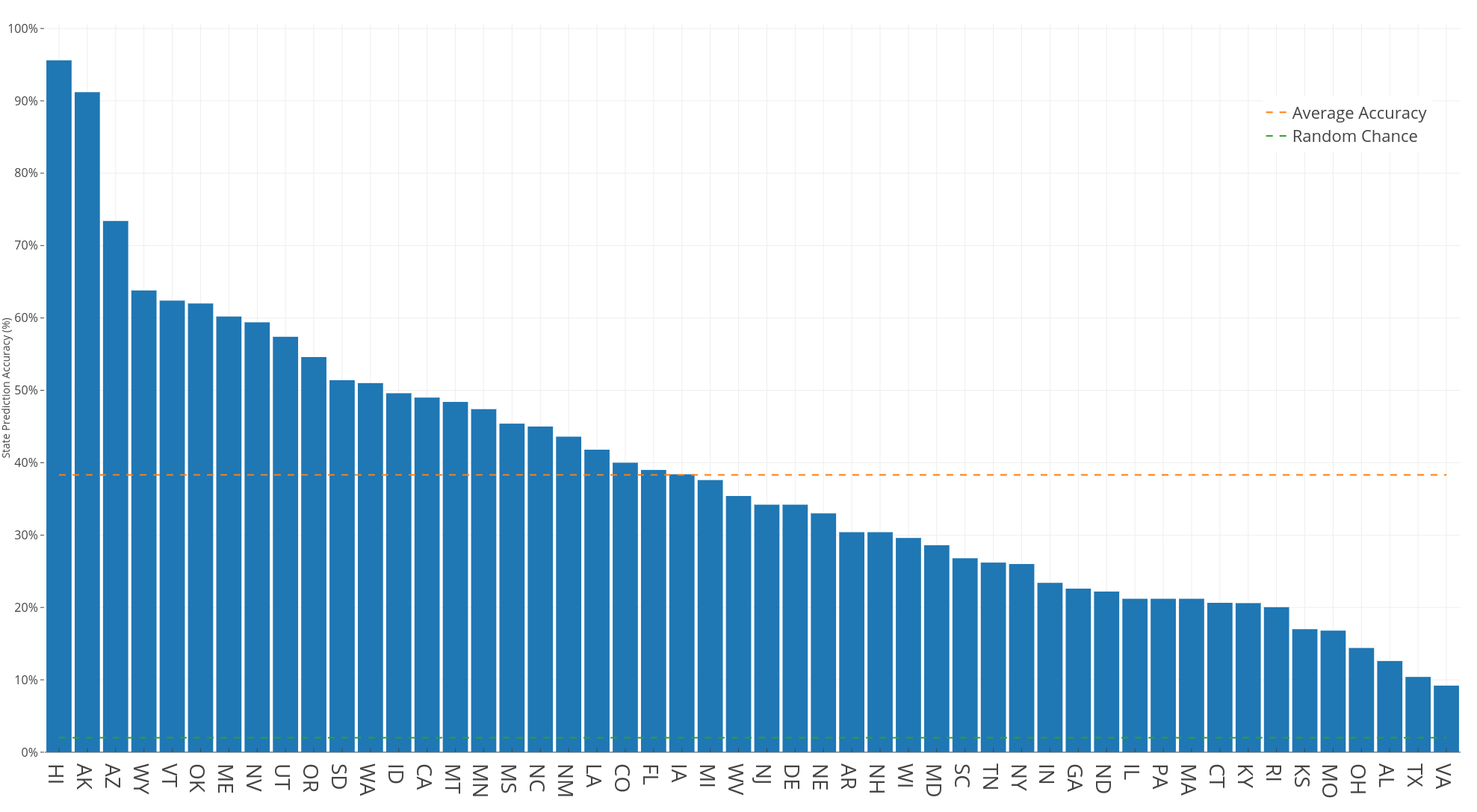}
 \caption{Accuracy among the 50 classes with the medium integration model on \texttt{50State2K} (best viewed on computer screen).}
 \label{fig:class_acc}
\end{figure*}

\section{Experiments and Results}  \label{results}

\subsection{Against Test Dataset}  \label{testresults}

Accuracy results for the test dataset \textbf{50States2K} amongst the four different network architectures are shown below. Figure \ref{fig:graph} depicts the average classification accuracy on the test dataset. For each architecture, we calculate the performance of the best in top-N predictions for an image. For this we sort the probability distribution of a given test image and check if the any of the greatest N values belong to the ground truth class. The rationale behind computing this metric is that some states have geographical and ecological similarities (irrespective of their proximity to each other). These are best visualized in the heatmap generated for images from Hawaii and Vermont (Figure \ref{fig:heatmaps}). 

\begin{figure}[t]
 \centering
 \includegraphics[width=\linewidth]{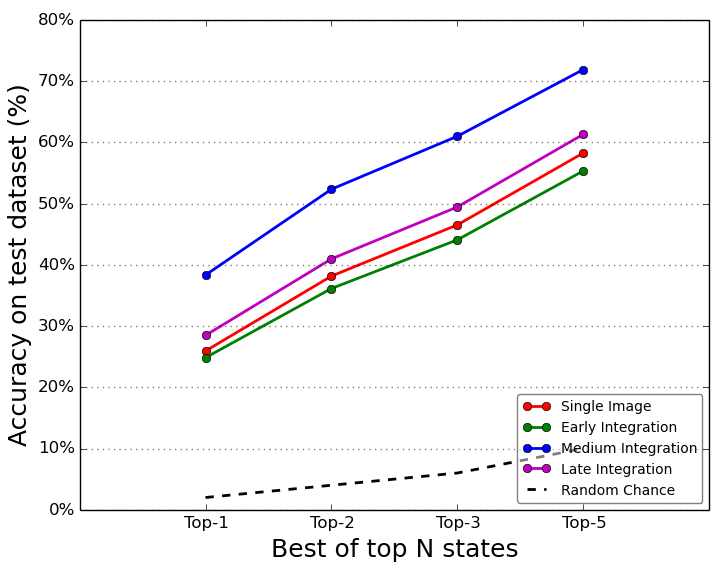}
 \caption{Best in top-N accuracy for different network architectures, compared with random chance.}
 \label{fig:graph}
\end{figure}

\begin{figure}[!t]
\captionsetup{width=\columnwidth}
 \centering
 \includegraphics[width=\columnwidth]{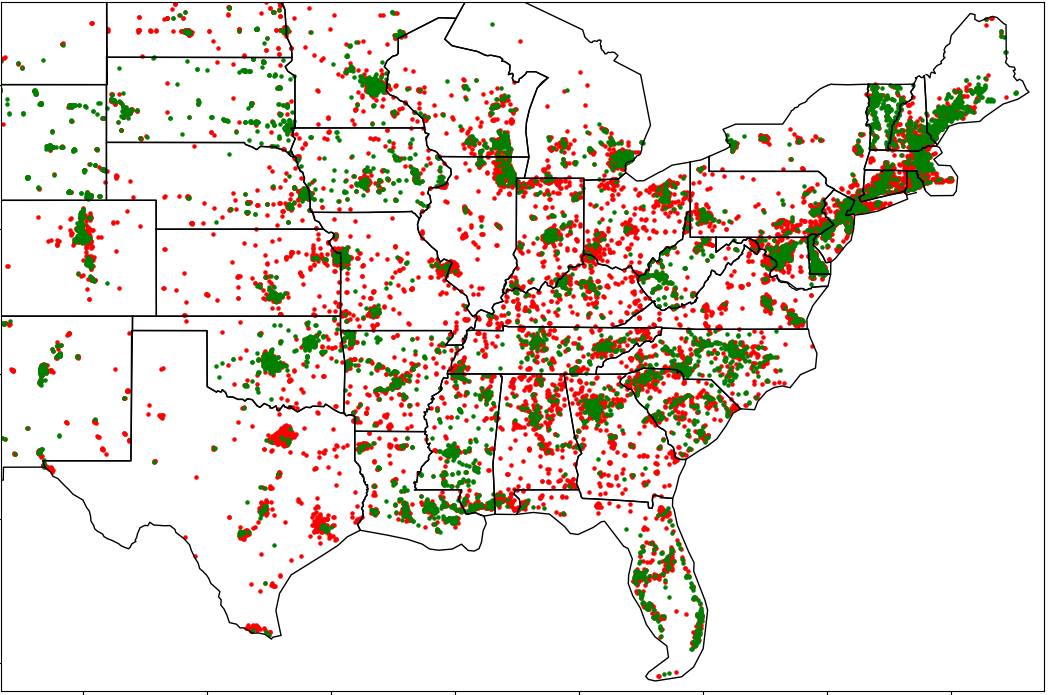}
 \caption{Test locations depicted as correctly/incorrectly classified. This show that the network performs poorly on samples from regions of low population density.}
 \label{fig:correct}
\end{figure} 

\begin{table}[h]
\centering
\label{accuracy}
\renewcommand{\arraystretch}{1.2}
\scalebox{0.9}{
\begin{tabular}{lcccc}
\Xhline{3\arrayrulewidth}
\multicolumn{1}{c}{Method} & \textbf{Top 1} & \textbf{Top 2} & \textbf{Top 3} & \textbf{Top 5} \\ \Xhline{3\arrayrulewidth} 
\textit{Single Image} & 25.92 & 38.15 & 46.51 & 58.25 \\
\textit{Early Integration} & 24.83 & 36.12 & 44.07 & 55.32 \\
\rowcolor{yellow} \textit{Medium Integration} & 38.32 & 52.33 & 60.98 & 71.87 \\
\textit{Late Integration} & 28.47 & 40.96 & 49.44 & 61.30\\
\Xhline{3\arrayrulewidth}
\end{tabular}}
\caption{Overall \% accuracy of each network architecture on the test dataset \textbf{50State2K}.}
\end{table}

We see that even in the single image case, a baseline accuracy of 25.92\% is obtained. The Medium Integration method yields the best results, with a 38.32\% accuracy for the top-1 metric. This confirms our initial intuition that performing feature extraction on the individual cardinal directions before integration is the preferred architecture. Interestingly, the Early Integration method performs poorly, even when compared to the Single Image method. 

\begin{figure*}[t]
 \centering
 \includegraphics[width=0.9\linewidth]{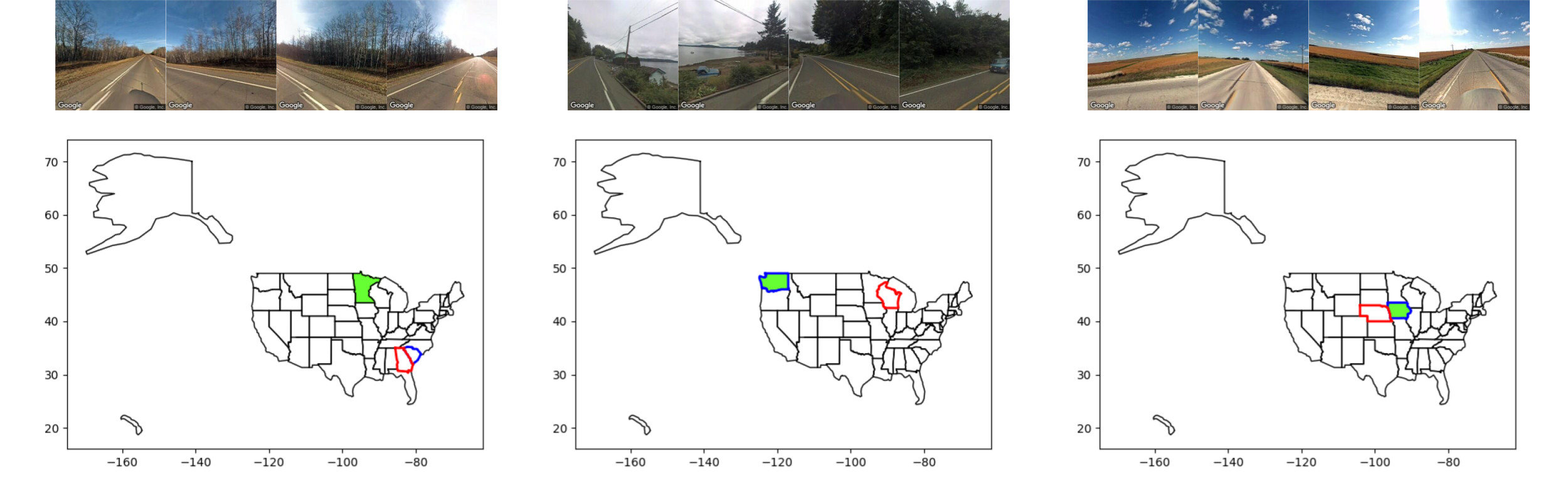}
 \caption{Correct label (\textcolor{green}{green}), human guess (\textcolor{red}{red}) and DeepGeo guess (\textcolor{blue}{blue}) displayed for a few human test cases.}
 \label{fig:mapComparison}
\end{figure*}

Across the board, the accuracies improve significantly when considering the best in top-N results. This peaks at 71.87\% for the Medium Integration  method. PlaNet correctly classifies 28.4\% of the test data at the country level. This cannot be considered to be a prior baseline for our paper, for a host of different reasons, expanded in section \ref{rel}. Random chance prediction for each image are 1-in-50 or 2\%. For the subsequent results, we consider those from the best-performing Medium Integration architecture. 

\begin{table}[!t]
\centering
\scalebox{0.9}{
\begin{tabular}{cc|cc}
\Xhline{3\arrayrulewidth}
\multicolumn{2}{c|}{\textbf{Best 5}} & \multicolumn{2}{c}{\textbf{Bottom 5}} \\ 
\textbf{State}       & \textbf{Accuracy (\%)}       & \textbf{State}              & \textbf{Accuracy (\%)}           \\ 
\Xhline{3\arrayrulewidth}
\textit{Hawaii}                & 95.6                         & \textit{Virginia}           & 9.2                              \\ 
\textit{Alaska}                & 91.2                         & \textit{Texas}              & 10.4                             \\ 
\textit{Arizona}               & 73.4                         & \textit{Alabama}            & 12.6                             \\ 
\textit{Wyoming}               & 63.8                         & \textit{Ohio}               & 14.4                             \\ 
\textit{Vermont}              & 62.4                         & \textit{Missouri}           & 16.8                             \\ 
\Xhline{3\arrayrulewidth}
\end{tabular}}
\caption{Best and worst performing classes by \% accuracy}
\label{best_worst}
\end{table}

We also analyze the per-class accuracies, to identify \textit{distinctness} trends amongst states. For example, its natural to assume that a model would be able to predict images in states like Hawaii and Alaska far better than those in the contiguous USA. This is due to their disparate landscape in comparison with other states. Table \ref{best_worst} presents the best and worst performing states, which falls in line with our assumption. Figure \ref{fig:class_acc} arranges the class accuracies of all 50 states. 

Figure \ref{fig:correct} plots correctly and incorrectly classified test samples. We observe that the network performs poorly on regions of low population density. As we sample based on population density due to the distribution of Street View images, the classifier is not well-trained on image features in these regions. 

\subsection{Against Humans in Geoguessr}  \label{humanresults}
The performance of our network was evaluated in head-to-head rounds of GeoGuessr against human subjects. 10 locations tests were performed per game. To ensure a fair test, humans were only allowed to rotate the panorama viewpoint and could not explore the map. At each location, coordinates were logged and used to download images facing the cardinal directions for network testing. It was necessary to replay roughly 50\% of the tests (with a different human player) as some locations were rejected by the Streetview API downloader. Table \ref{tab:human_trials} shows the results of these tests. The medium integration network beat humans in 4 out of 5 rounds. 

Figure \ref{fig:mapComparison} shows representative results from 3 rounds. The leftmost image shows that mis-classifications typically arise when samples strongly resemble locations in different states. In this case, both humans and DeepGeo place guesses in the wrong region. The next two demonstrate DeepGeo outperforming the human in both large and small scale accuracy.

\begin{table}[t]
\captionsetup{width=\columnwidth}
\centering
\scalebox{0.8}{
\begin{tabular}{ccccc}
\Xhline{3\arrayrulewidth}
Game & \textbf{Human (\%)} & \textbf{Early (\%)} & \textbf{Medium (\%)} & \textbf{Late (\%)} \\ \Xhline{3\arrayrulewidth}
$1$ & $0$ & $10$ & $60$ & $20$\\ 
$2$ & $0$ & $10$ & $10$ & $0$\\ 
$3$ & $30$ & $10$ & $10$ & $10$\\ 
$4$ & $10$ & $30$ & $30$ & $10$\\ 
$5$ & $10$ & $10$ & $20$ & $0$\\ 
\Xhline{3\arrayrulewidth}
\end{tabular}}
\caption{Accuracy (\%) over 10 rounds of \textit{GeoGuessr} for human subject and early, medium and late integration architectures.}
\label{tab:human_trials}
\end{table}

\section{Conclusion and Future Work}

GeoGuessr poses a challenging image classification task of estimation location based on image pixels. We have presented several DeepGeo network architectures for this task, achieving excellent performance with the best variant. In addition, we open-sourced both a training and test dataset for similar tasks. We explored importance of choosing the appropriate information integration strategy for a classification task. Specifically, we have shown that in contrast to conventional wisdom, the earliest integration scheme is not always superior. 

For future work, we would like to extend the model to perform regression on latitude and longitude rather than classification on state. This would also include collecting a uniformly sampled dataset, as opposed to population-based sampling. We also wish to evaluate results with different network architectures - for example, PlaNet uses the Inception architecture to obtain good results. The ideal use-case we would like to see, is when the network is able to generalize to in-the-wild photographs that don't belong to the street view dataset.

\vfill
\bibliographystyle{apalike}
{\small
\bibliography{example}}

\begin{thebibliography}{}

\bibitem[Arandjelovic et~al., 2016]{arandjelovic2016netvlad}
Arandjelovic, R., Gronat, P., Torii, A., Pajdla, T., and Sivic, J. (2016).
\newblock {N}et{VLAD}: {CNN} architecture for weakly supervised place
  recognition.
\newblock In {\em Proceedings of the IEEE Conference on Computer Vision and
  Pattern Recognition}, pages 5297--5307.

\bibitem[Cao and Snavely, 2013]{cao2013graph}
Cao, S. and Snavely, N. (2013).
\newblock Graph-based discriminative learning for location recognition.
\newblock In {\em Computer Vision and Pattern Recognition (CVPR), 2013 IEEE
  Conference on}, pages 700--707. IEEE.

\bibitem[Hays and Efros, 2008]{hays2008im2gps}
Hays, J. and Efros, A.~A. (2008).
\newblock {IM2GPS}: estimating geographic information from a single image.
\newblock In {\em Computer Vision and Pattern Recognition, 2008. CVPR 2008.
  IEEE Conference on}, pages 1--8. IEEE.

\bibitem[He et~al., 2015]{resnet}
He, K., Zhang, X., Ren, S., and Sun, J. (2015).
\newblock Deep residual learning for image recognition.
\newblock {\em CoRR}, abs/1512.03385.

\bibitem[Kim et~al., ]{kim2015predicting}
Kim, H.~J., Dunn, E., and Frahm, J.-M.
\newblock Predicting good features for image geo-localization using per-bundle
  {VLAD}.

\bibitem[Ramalingam et~al., 2010]{ramalingam2010skyline2gps}
Ramalingam, S., Bouaziz, S., Sturm, P., and Brand, M. (2010).
\newblock Skyline2{GPS}: Localization in urban canyons using omni-skylines.
\newblock In {\em Intelligent Robots and Systems (IROS), 2010 IEEE/RSJ
  International Conference on}, pages 3816--3823. IEEE.

\bibitem[Thompson et~al., 1996]{thompson1996geometric}
Thompson, W.~B., Valiquette, C.~M., Bennet, B., and Sutherland, K.~T. (1996).
\newblock Geometric reasoning for map-based localization.
\newblock {\em Computer Science Technical Report UUCS-96-006, University of
  Utah, Salt Lake City, UT}.

\bibitem[Weyand et~al., 2016]{weyand2016planet}
Weyand, T., Kostrikov, I., and Philbin, J. (2016).
\newblock {P}la{N}et-photo geolocation with convolutional neural networks.
\newblock In {\em European Conference on Computer Vision}, pages 37--55.
  Springer.

\bibitem[Zamir and Shah, 2010]{zamir2010accurate}
Zamir, A.~R. and Shah, M. (2010).
\newblock Accurate image localization based on google maps street view.
\newblock In {\em European Conference on Computer Vision}, pages 255--268.
  Springer.

\bibitem[Zamir and Shah, 2014]{zamir2014image}
Zamir, A.~R. and Shah, M. (2014).
\newblock Image geo-localization based on multiple nearest neighbor feature
  matching using generalized graphs.
\newblock {\em IEEE transactions on pattern analysis and machine intelligence},
  36(8):1546--1558.

\end{thebibliography}
\end{document}